\documentclass[a4paper]{article}
\usepackage{MM805}
\usepackage{adjustbox}
\usepackage{tikz,graphics,color,fullpage,float,epsf,caption,subcaption}
\usepackage{listings}
\usepackage{tikz,lipsum,lmodern}
\usepackage{algorithm}

\usepackage[most]{tcolorbox}
\usepackage{algcompatible}
\author{\begin{tabular}{ccc}
Md Alif Rahman Ridoy & M Mahmud Hasan & Shovon Bhowmick \\
University of Alberta & University of Alberta & University of Alberta \\
\textit{mdalifra@ualberta.ca} & \textit{mmahmud3@ualberta.ca} & \textit{sbhowmic@ualberta.ca} \\
\end{tabular}}

\title{Compressed Image Captioning using CNN-based Encoder-Decoder Framework}

\begin{document}
\maketitle
\section*{Abstract}
In today's world, image processing plays a crucial role across various fields, from scientific research to industrial applications. But one particularly exciting application is image captioning. The potential impact of effective image captioning is vast. It can significantly boost the accuracy of search engines, making it easier to find relevant information. Moreover, it can greatly enhance accessibility for visually impaired individuals, providing them with a more immersive experience of digital content. However, despite its promise, image captioning presents several challenges. One major hurdle is extracting meaningful visual information from images and transforming it into coherent language. This requires bridging the gap between the visual and linguistic domains, a task that demands sophisticated algorithms and models. Our project is focused on addressing these challenges by developing an automatic image captioning architecture that combines the strengths of convolutional neural networks (CNNs) and encoder-decoder models. The CNN model is used to extract the visual features from images, and later, with the help of the encoder-decoder framework, captions are generated. We also did a performance comparison where we delved into the realm of pre-trained CNN models, experimenting with multiple architectures to understand their performance variations. In our quest for optimization, we also explored the integration of frequency regularization techniques to compress the "AlexNet" and "EfficientNetB0" model. We aimed to see if this compressed model could maintain its effectiveness in generating image captions while being more resource-efficient.

\section{Introduction}
In recent times, the blending of computer vision and natural language processing has brought forth an intriguing area called image captioning. This field isn't just about smart thinking; it's like a deep dive into how machines can see and talk like humans. Basically, image captioning is about computers describing pictures in words all by themselves. It's like giving them the ability to understand and explain what they see in images. And this isn't just important for one thing—it's got a ton of uses in all sorts of areas.

\noindent
Image captioning is pivotal in developing assistive technologies that cater to the needs of visually impaired individuals. By providing descriptive captions for images, it enables them to access digital content more comprehensively and independently\cite{dognin2022image}. On social media platforms, image captioning enhances accessibility and engagement by making visual content more accessible to users. It enriches the browsing experience by providing informative descriptions alongside images, facilitating greater understanding and interaction \cite{monteiro2017situational}. In content moderation, image captioning aids in identifying and flagging inappropriate or harmful content. By analyzing images and generating descriptive captions, it assists moderators in identifying content that violates community guidelines or standards \cite{gerrard2018beyond}. Image captioning plays a crucial role in medical imaging by providing descriptive annotations for medical images. It aids healthcare professionals in interpreting and analyzing complex medical images, leading to more accurate diagnosis and treatment planning \cite{park2021medical}. In automated surveillance systems, image captioning enhances situational awareness by providing contextual information about detected events or activities. It enables automated systems to generate descriptive captions for surveillance images, facilitating quicker response and decision-making. Image captioning powers visual search engines by enabling them to understand and interpret visual content. It facilitates more accurate and relevant search results by providing descriptive captions for images, enhancing the user experience \cite{goyal2023captionomaly}. In robotics, image captioning enables robots to perceive and understand their environment more effectively. It provides descriptive annotations for visual scenes, enabling robots to navigate and interact with their surroundings autonomously \cite{luo2019visual}. It enhances educational content by providing descriptive annotations for visual materials. It enables educators to create more accessible and engaging learning resources, facilitating comprehension and retention. Image captioning automates the process of image tagging by generating descriptive captions for images. It assists in organizing and categorizing large collections of images, improving searchability and accessibility \cite{hasnine2019vocabulary}. It also enables the creation of descriptive narratives for virtual scenes and objects in virtual reality environments. It enhances immersion and interactivity by providing contextually relevant descriptions for virtual elements. Also, image captioning assists in environmental monitoring efforts by providing descriptive annotations for satellite or drone images. It aids in analyzing environmental changes and trends, facilitating informed decision-making and policy planning. Due to these implementations of image captioning, it has garnered the attention of researchers for quite some time now, leading to the development of numerous advanced algorithms aimed at enhancing its performance \cite{lu2021environment}.

\noindent
In our work, we started by utilizing a CNN model for feature extraction. Specifically, we employed several state-of-the-art pre-trained CNN models for this purpose. These models are well-known for their ability to extract rich and meaningful features from images. Once we obtained the extracted features, we fed them into an encoder-decoder framework to generate the final captions. This framework takes the visual features extracted by the CNN model and processes them through an encoder, which encodes the information in a way that is suitable for generating captions. Subsequently, a decoder takes this encoded information and generates descriptive captions based on it. By combining the power of CNNs for feature extraction with the flexibility of the encoder-decoder architecture for caption generation, we aimed to develop a robust and efficient image captioning system. 

\section{Literature Review}
This paper \cite{ghandi2023deep}highlights the importance of image captioning and the transformative impact of deep learning and vision-language pre-training techniques on the field. It talks about various challenges in image captioning, such as object hallucination, missing context, illumination conditions, contextual understanding, and referring expressions. This article \cite{sairam2021image} proposes a model for image caption generation using deep neural networks. The model takes an image as input and generates output in the form of a sentence in three different languages, an mp3 audio file, and an image file. The model combines computer vision and natural language processing techniques, specifically utilizing convolutional neural networks (CNN) and long short-term memory (LSTM) to generate captions. The model compares the target image with a large dataset of training images using CNN and then utilizes LSTM to decode the description of the image. The author of this paper \cite{Aneja_2018_CVPR} explores the task of image captioning and highlights its importance across various applications, such as virtual assistants, editing tools, image indexing, and support for the disabled. The authors also talk about the significant progress made in image captioning using recurrent neural networks (RNNs) powered by long-short-term memory (LSTM) units. Though the LSTM units have advantages in lowering the vanishing gradient problem and memorizing dependencies, they are known to be complex indicating a potential limitation in their performance. In response to this limitation, the paper introduces a convolutional approach for image captioning. This approach was developed to demonstrate the benefits of convolutional networks for machine translation and conditional image generation in previous research.The paper \cite{devlin2015language} compares two recent approaches to image captioning that have achieved state-of-the-art results. The first approach involves using a convolutional neural network (CNN) to generate a set of candidate words, followed by arranging these words into a coherent sentence using a maximum entropy (ME) language model. The second approach utilizes the penultimate activation layer of the CNN as input to a recurrent neural network (RNN) to generate the caption sequence. The authors compare the merits of these language modeling approaches for the first time, using the same state-of-the-art CNN as input. They address issues such as linguistic irregularities, caption repetition, and dataset overlap. The authors of this paper \cite{lebret2015phrase} propose a model for generating descriptive sentences based on images, bridging the gap between computer vision and natural language processing. The model emphasizes the syntax of the descriptions and utilizes a purely bi-linear model trained to learn a metric between image representations and corresponding phrases. It leverages a convolutional neural network to generate image representations. The system can then infer descriptive phrases from an input image and utilize a language model based on caption syntax statistics to produce relevant descriptions. Despite its simplicity compared to existing models, the approach yields comparable results in popular datasets like Flickr 30k and Microsoft COCO. The paper proposes \cite{cho2015describing} a novel approach for generating textual descriptions of multimedia content, such as images and videos, using attention-based encoder-decoder networks. The encoder-decoder architecture is a neural network framework commonly used for tasks involving sequence-to-sequence learning, where an input sequence is mapped to an output sequence. In this paper, the authors extend this framework by incorporating an attention mechanism, which allows the model to focus on specific parts of the input sequence when generating the output sequence. The proposed approach consists of two main components: an encoder network and a decoder network. The encoder network processes the input multimedia content and produces a fixed-length representation, or "encoding," which captures the semantic information of the content. The decoder network then takes this encoding as input and generates a textual description of the content.  In this paper \cite{farhadi2010every}, the authors designed a system to generate concise descriptive sentences from images like humans.  The procedure is calculating a score that connects an image to a statement. This score can be used to identify photos that support a particular text or to append a descriptive sentence to an existing image. The paper outlines the approach to mapping images to meaning, which involves learning the mapping from images to meaning discriminatively using pairs of images and assigned meaning representations. It also delves into the methodology for predicting triplets for images, with a focus on object, action, and scene classifications, and the evaluation of mapping images to meanings using measures such as Tree-F1 and BLUE. In this paper \cite{vinyals2015show}, the authors proposed a model to describe an image using a generative model based on a deep recurrent architecture that combines computer vision and machine translation. The key components of the model include a vision Convolutional Neural Network (CNN) followed by a language generating Recurrent Neural Network (RNN). This end-to-end neural network system processes an input image through the CNN to extract visual features, which are then fed into the RNN to generate complete sentences in natural language. This seamless integration of visual understanding and language generation enables the model to capture not only the objects in the image but also their relationships, attributes, and activities. To address challenges such as overfitting, the model leverages techniques like weight initialization from pre-trained models, particularly on large datasets like ImageNet. Despite the relatively small training datasets used in the experiments, the model demonstrates promising results, with accuracy and fluency verified both qualitatively and quantitatively. This model beats the state-of-the-art models in BLEU-1 score on the Pascal dataset over a huge margin. This model yields a score of 59 compared to 25, scored by the best model available previously.  In this paper \cite{wang2016cnnpack} the authors first address the challenges of storage and computational requirements, particularly for mobile devices. Then proposes an approach to compress not only smaller weights but all weights and their underlying connections. By leveraging the frequency domain and the Discrete Cosine Transform (DCT), the proposed algorithm aims to achieve high compression ratios and speed gains compared to existing methods. The representation of convolutional filters in the frequency domain is decomposed into common parts shared by similar filters and private parts containing unique information. By discarding subtle frequency coefficients, both parts can be compressed significantly. An efficient convolution calculation scheme is developed to exploit the relationship between feature maps of DCT bases and frequency coefficients, leading to theoretical discussions on compression and speed-up. By exploring a cheaper convolution calculation based on the sparsity of the compressed network in the frequency domain, CNNpack maintains the functionality of the original network while achieving high compression and speed-up ratios. In the paper named "DepGraph: Towards Any Structural Pruning" \cite{fang2023depgraph}, the authors proposed the idea of  “Any structural pruning”, to tackle general structural pruning of arbitrary architecture like CNNs, RNNs, GNNs, and Transformers. The most difficult task of this architecture was structural coupling. To avoid this, the authors proposed Dependency Graph (DepGraph) to explicitly model the dependency between layers and group-coupled parameters for pruning. The authors demonstrated that with a simple norm-based criterion, the proposed method showed good results. In this paper \cite{zhao2023frequency}, the authors aim to limit the network parameters in the frequency domain by applying Frequency Regularization. As the architecture operates at a tensor level, it can be adapted to state-of-the-art architectures like LeNet, Alexnet, VGG, Resnet, etc. The human eye mainly perceives the low-frequency components and often ignores high-frequency components, this architecture is built on this phenomenon. LeNet5 with 0.4M parameters can be represented by only 776 float16 numbers. As the features learned by the neural networks are closely related to the training images it is expected that they would mirror the same trend like high-frequency components are less important. However small changes in the key parameters can degrade the performance of the network. For this reason, the authors focused on using the frequency domain transform as a network regulator during the training process. Although frequency regularizations are used to prevent overfitting, this paper extended the use case to restrict the number of non-zero parameters. The tensors of network parameters are maintained in the frequency domain instead of the spatial domain, This helps to truncate the tail elements of the tensors by zigzag mask matrix. The frequency tensors are again fed to the IDCT (Inverse Discrete Cosine Transform) to reconstruct the spatial tensors. Then the training is done using the reconstructed spatial tensors. As the remaining parameters may not have suitable gradients for backpropagation, the authors proposed a dynamic trail-truncation strategy. Here, as the total number of truncated parameters increases, the number of parameters truncated in each training epoch decreases.

\section{Proposed Methodology}

In our methodology, we first pass the image data through a pre-trained CNN model to extract features. These features capture important information about the images. Next, we feed these features into a transformer encoder model. This encoder model transforms the features into a new representation that helps the model understand the images better. Then, we combine this new representation with the ground truth captions and feed them into a decoder model. This decoder model predicts the final output caption based on the combined information

\begin{figure}[h]
\centering
\includegraphics[height=2cm, width=0.7\textwidth]{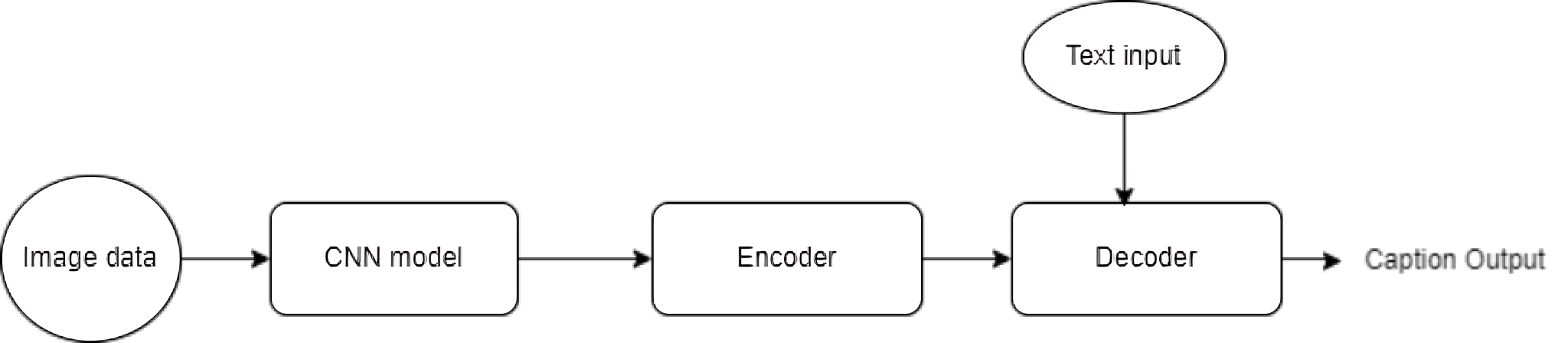}
\caption{Proposed System}
\end{figure}

\subsection{Dataset}
We utilized the 2014 Train/Val COCO dataset for our project \cite{COCOdataset}. There is a total of 82783 train images and 40504 validation images in the original dataset. In our work, we first processed the data and made it suitable for our purpose. Here are some samples of the dataset.

\begin{figure}[h]
     \centering
     \begin{subfigure}[b]{0.27\textwidth}
         \centering
         \includegraphics[width=\textwidth, height = 3 cm]{Figures/COCO_val2014_000000179765.jpg}
         \caption{}
         \label{}
     \end{subfigure}
     \hfill
     \begin{subfigure}[b]{0.27\textwidth}
         \centering
         \includegraphics[width=\textwidth, height = 3 cm]{Figures/COCO_val2014_000000203564.jpg}
         \caption{}
         \label{}
     \end{subfigure}
     \hfill
     \begin{subfigure}[b]{0.27\textwidth}
         \centering
         \includegraphics[width=\textwidth, height = 3 cm]{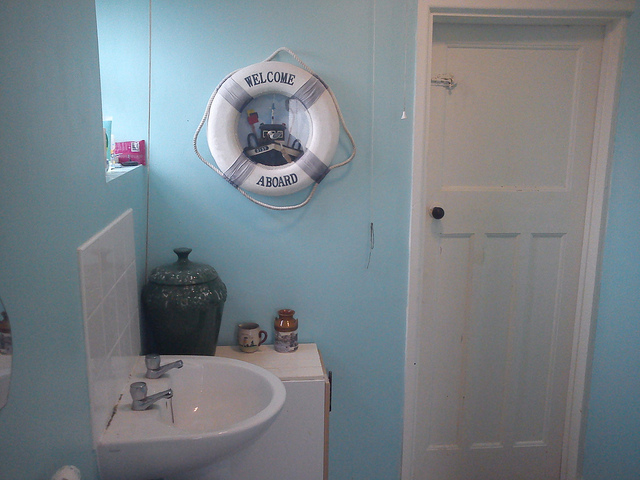}
         \caption{}
         \label{}
     \end{subfigure}
        \caption{Data Sample}
        \label{fig:three graphs}
\end{figure}

\subsection{Dataset Preprocessing}
The initial dataset comprises 82,783 training images and 40,504 validation images, with each image having between 1 and 6 captions. To streamline the dataset, we conducted preprocessing to retain only images with exactly 5 captions. This preprocessing step ensured that the model was trained using a consistent set of 5 captions for each image. Following this filtering process, the revised dataset consists of 68,363 training images, 31,432 validation images and 2000 testing images. For training, validating and testing three different \textit{json} files were also prepared. The json files contain image paths and it's corresponding captions. Here is the structure of our prepared file \textit{"path\_to\_image1.jpg": ["caption1", "caption2", "caption3", "caption4", "caption5"], "path\_to\_image2.jpg": ["caption1", "caption2", "caption3", "caption4", "caption5"], ..}. 

\begin{figure}[h]
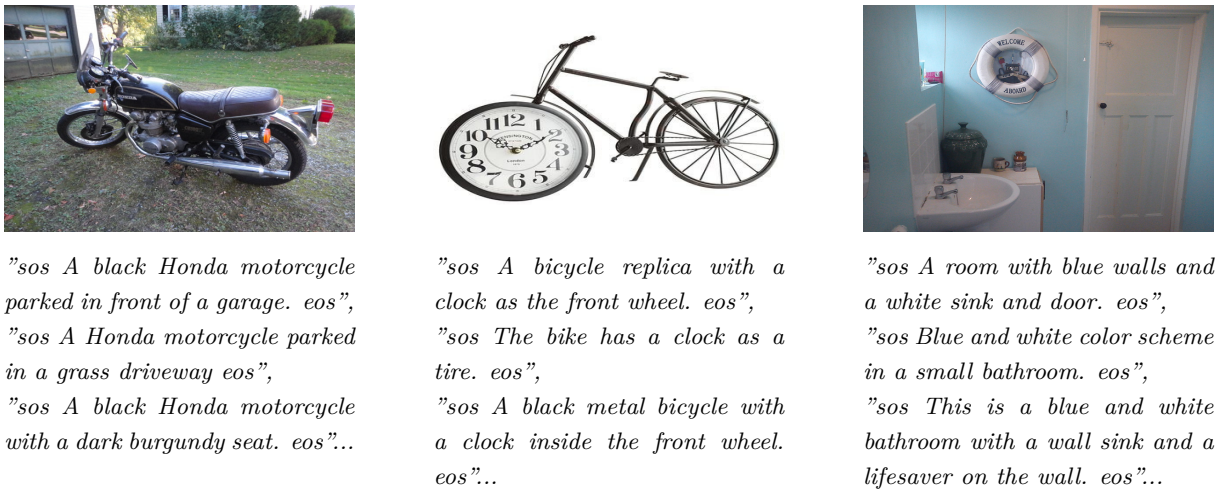

     \centering
     \begin{subfigure}[t]{0.29\textwidth}
        \centering
        \includegraphics[width=\textwidth, height = 3 cm]{Figures/COCO_val2014_000000179765.jpg}
        \caption*{\small{\textit{"sos A black Honda motorcycle parked in front of a garage. eos",\\
        "sos A Honda motorcycle parked in a grass driveway eos",\\
        "sos A black Honda motorcycle with a dark burgundy seat. eos"...}}}
    \end{subfigure}
     \hfill
     \begin{subfigure}[t]{0.29\textwidth}
         \centering
         \includegraphics[width=\textwidth, height = 3 cm]{Figures/COCO_val2014_000000203564.jpg}
         \caption*{\small{\textit{"sos A bicycle replica with a clock as the front wheel. eos",\\
        "sos The bike has a clock as a tire. eos",\\
        "sos A black metal bicycle with a clock inside the front wheel. eos"...}}}
         \label{}
     \end{subfigure}
     \hfill
     \begin{subfigure}[t]{0.29\textwidth}
         \centering
         \includegraphics[width=\textwidth, height = 3 cm]{Figures/COCO_val2014_000000322141.jpg}
         \caption*{\small{\textit{"sos A room with blue walls and a white sink and door. eos",\\
        "sos Blue and white color scheme in a small bathroom. eos",\\
        "sos This is a blue and white bathroom with a wall sink and a lifesaver on the wall. eos"...}}}
         \label{}
     \end{subfigure}
        \caption{Data Sample with captions}
        \label{fig:three graphs with captions}
\end{figure}

\subsection{Training}
The first step in our training process is the feature extraction, performed by some of the state-of-the-art pre-trained CNN models. We used four different models, "EfficientNetB0", "EfficientNetB1", "ResNet50", and "MobileNetV2" which are pre-trained on the “imagenet” dataset. First, we loaded the pre-trained weights by importing with the help of \textit{keras} library. The feature extraction layers were made frozen so that they don't get trained. This was done so that the original weights are preserved and get used to perform the feature extraction on our dataset. 

\noindent
Data augmentation was performed to make the learning more generalized. We used below data augmentation parameters in our training process.
\\

\begin{algorithm}
\caption{Data Augmentation}
\begin{algorithmic}[1]
\STATE Initialize \textit{trainAug} as a Sequential model
\STATE Add Random Contrast layer to \textit{trainAug} with factor=0.05
\STATE Add Random Translation layer to \textit{trainAug} with height\_factor=0.1 and width\_factor=0.1
\STATE Add Random Zoom layer to \textit{trainAug} with height\_factor=(-0.1, 0.1) and width\_factor=(-0.1, 0.1)
\STATE Add Random Rotation layer to \textit{trainAug} with factor=(-0.1, 0.1)
\end{algorithmic}
\end{algorithm}
\vspace{0.5cm}

\noindent
The data was placed under required folders to start the training process. Different training parameters are set like "IMAGE\_SIZE = (299, 299), MAX\_VOCAB\_SIZE = 2000000, SEQ\_LENGTH = 25, EMBED\_DIM = 512, NUM\_HEADS = 6, FF\_DIM = 1024, SHUFFLE\_DIM = 256, BATCH\_SIZE = 16, EPOCHS = 14, NUM\_TRAIN\_IMG = 68363, NUM\_VALID\_IMG = 31432, TRAIN\_SET\_AUG = True, VALID\_SET\_AUG = False, TEST\_SET = True".

\subsection{Performance Evaluation}

Here is a comparison table for all models with different CNN pre-trained weights. For performing prediction we used two different ways. First, if an image path is passed as argument then the trained model will perform prediction on that image only and generate output in console. Secondly, if no path is passed, then the model will perform prediction on each of the images form the \textit{test.json} file, which is our testing dataset. Each prediction along with the corresponding images will be stored in a new \textit{prediction.json} file. This two \textit{json} files are later used for calculating the other evaluation metrics. As for the evaluation metrics, we used three different metrics. Each image has 5 corresponding captions as ground truth in the \textit{test.json} file. While calculating the metric values, the predicted output was evaluated against all the 5 captions and a average was taken as the final value. For BLEU, 4 different metric is used, named as BLEU-1,2,3 and 4. These represents the percentage of similarities for consecutive 1,2,3 and 4 words between the actual and predicted output. Same thing goes for the ROUGE metric. As per the METEOR, it's just a single calculated value. 


\begin{table}[h]
\caption{Model Performance Comparison}
\begin{adjustbox}{width=\columnwidth,center}
\begin{tabular}{|l|l|l|l|l|l|l|l|l|l|}
\hline
\textbf{Model Name} & \textbf{\begin{tabular}[c]{@{}l@{}}Training\\ Accuracy\end{tabular}} & \textbf{\begin{tabular}[c]{@{}l@{}}Validation\\ Accuracy\end{tabular}} & \textbf{\begin{tabular}[c]{@{}l@{}}Testing\\ Accuracy\end{tabular}} & \textbf{\begin{tabular}[c]{@{}l@{}}Training\\ Loss\end{tabular}} & \textbf{\begin{tabular}[c]{@{}l@{}}Validation\\ Loss\end{tabular}} & \textbf{\begin{tabular}[c]{@{}l@{}}Testing\\ Loss\end{tabular}} & \textbf{BLEU score}                                                                                                & \textbf{ROUGE score}                                                                                 & \textbf{\begin{tabular}[c]{@{}l@{}}METEOR\\ score\end{tabular}} \\ \hline
EfficientNetB0      & 0.5028                                                               & 0.4793                                                                 & 0.4879                                                              & 13.8323                                                          & 14.9792                                                            & 14.0197                                                         & \begin{tabular}[c]{@{}l@{}}BLEU-1: 0.2827\\ BLEU-2: 0.1325\\ BLEU-3: 0.0588\\ BLEU-4: 0.0266\end{tabular}          & \begin{tabular}[c]{@{}l@{}}ROUGE-1: 0.4027\\ ROUGE-2: 0.1469\\ ROUGE-L: 0.3609\end{tabular}          & 0.2661                                                          \\ \hline
\textbf{EfficientNetB1}      & \textbf{0.5160}                                                      & \textbf{0.5063}                                                        & \textbf{0.5328}                                                     & \textbf{13.6735}                                                 & \textbf{14.1245}                                                   & \textbf{12.8375}                                                & \textbf{\begin{tabular}[c]{@{}l@{}}BLEU-1: 0.2890\\ BLEU-2: 0.1404\\ BLEU-3: 0.0642\\ BLEU-4: 0.0286\end{tabular}} & \textbf{\begin{tabular}[c]{@{}l@{}}ROUGE-1: 0.4117\\ ROUGE-2: 0.1551\\ ROUGE-L: 0.3718\end{tabular}} & \textbf{0.2710}                                                 \\ \hline
ResNet50            & 0.4521                                                               & 0.4484                                                                 & 0.5057                                                              & 16.3699                                                          & 16.5786                                                            & 13.8547                                                         & \begin{tabular}[c]{@{}l@{}}BLEU-1: 0.2637\\ BLEU-2: 0.1217\\ BLEU-3: 0.0496\\ BLEU-4: 0.0207\end{tabular}          & \begin{tabular}[c]{@{}l@{}}ROUGE-1: 0.3765\\ ROUGE-2: 0.1308\\ ROUGE-L: 0.3423\end{tabular}          & 0.2437                                                          \\ \hline
MobileNetV2         & 0.4366                                                               & 0.4295                                                                 & 0.4294                                                              & 17.8343                                                          & 17.8764                                                            & 16.3674                                                         & \begin{tabular}[c]{@{}l@{}}BLEU-1: 0.2106\\ BLEU-2: 0.0640\\ BLEU-3: 0.0215\\ BLEU-4: 0.0090\end{tabular}          & \begin{tabular}[c]{@{}l@{}}ROUGE-1: 0.2903\\ ROUGE-2: 0.0628\\ ROUGE-L: 0.2606\end{tabular}          & 0.1794                                                          \\ \hline
\end{tabular}
\end{adjustbox}
\end{table}

\vspace{0.8 cm}
\noindent
The models were trained for 14 epochs with each of the pre-trained weights. Below is the accuracy comparison among the models.

\begin{figure}[h]
     \centering
     \begin{subfigure}[b]{0.45\textwidth}
         \centering
         \includegraphics[width=\textwidth, height = 5.2 cm]{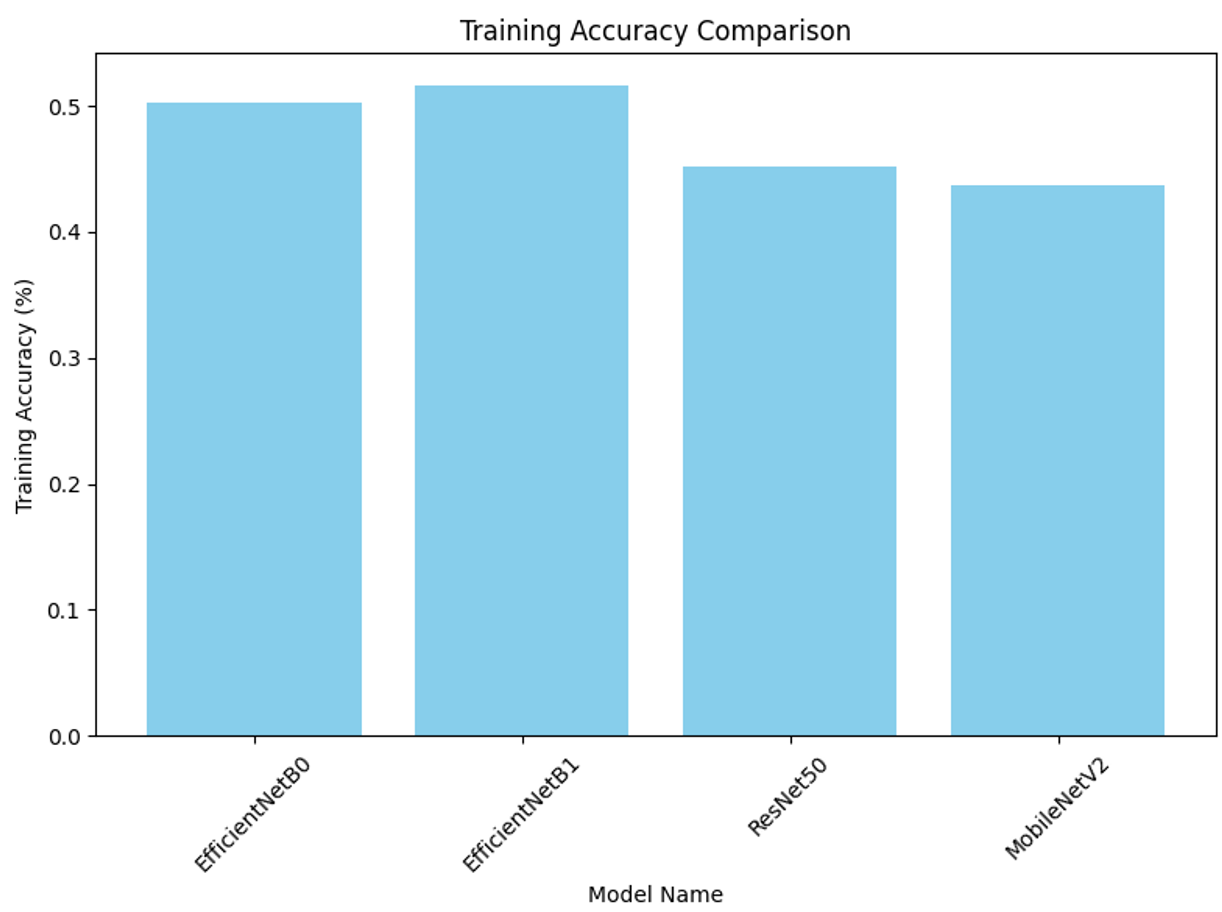}
         \caption{Training Accuracy comparison}
         \label{}
     \end{subfigure}
     \hfill
     \begin{subfigure}[b]{0.45\textwidth}
         \centering
         \includegraphics[width=\textwidth, height = 5.2 cm]{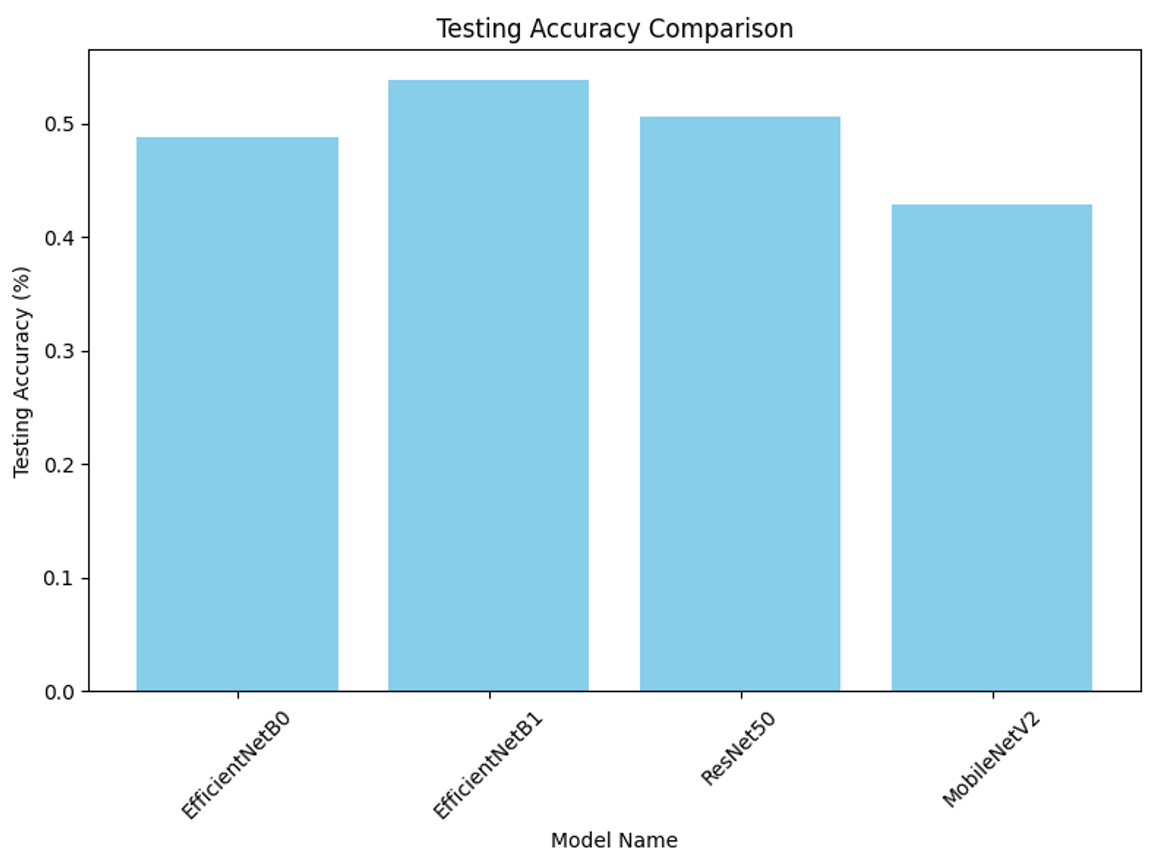}
         \caption{Testing Accuracy comparison}
         \label{}
     \end{subfigure}
        \caption{Accuracy comparison of models}
        \label{fig:three graphs}
\end{figure}


\vspace{1cm}

\noindent
These are the three image captioning evaluation metrics used in our work:

\begin{itemize}
    \item BLEU (Bilingual Evaluation Understudy): BLEU is a precision-based metric that measures the similarity between the generated captions and reference captions by counting the number of overlapping n-grams (contiguous sequences of n words) between them. BLEU ranges from 0 to 1, with higher scores indicating better similarity. It provides a quantitative measure of the quality of generated captions by assessing their similarity to human-generated reference captions \cite{9087226}. In our work we calculate BLEU-1,2,3 and 4 values and calculated an average from them. Below is the average value comparison.
    
\begin{figure}[h]
\centering
\includegraphics[height=4cm, width=0.4\textwidth]{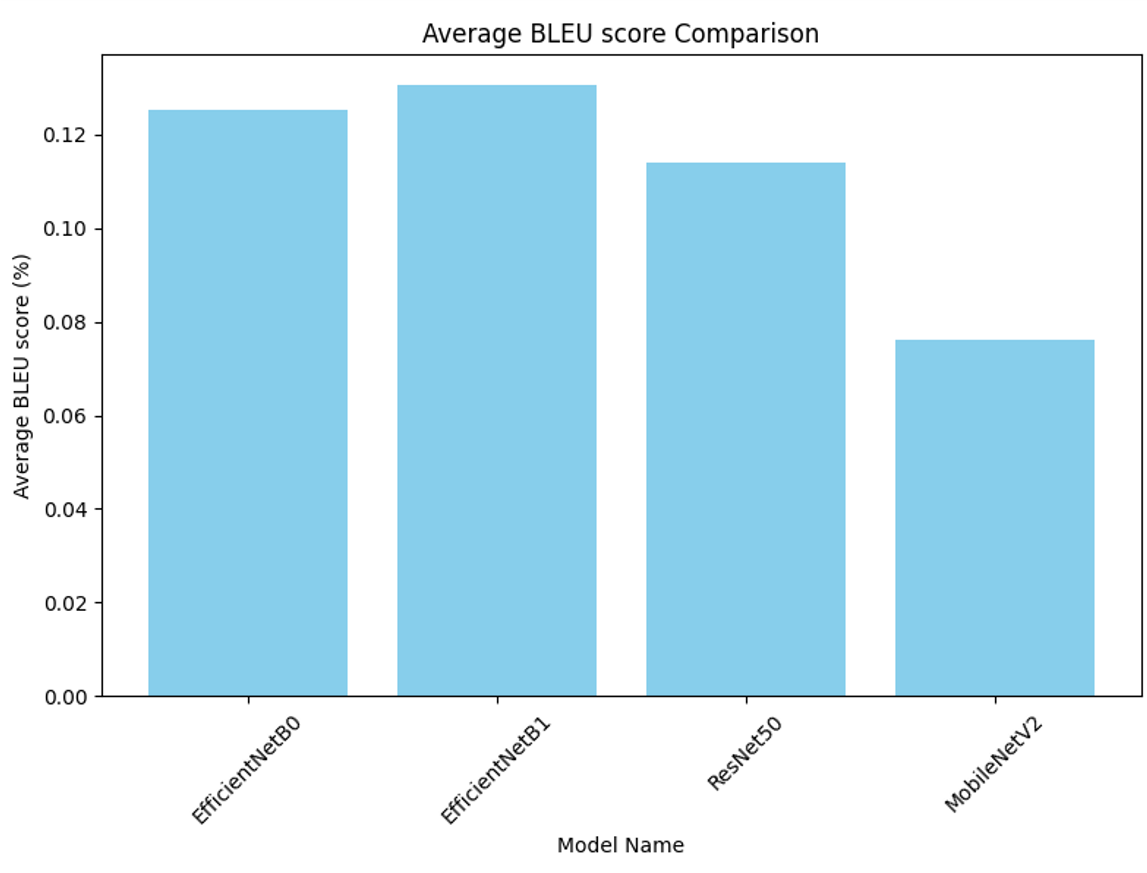}
\caption{Average BLEU score comparison}
\end{figure}

    \item ROUGE (Recall-Oriented Understudy for Gisting Evaluation): ROUGE is a recall-based metric that measures the overlap between the generated captions and reference captions using various measures such as n-gram overlap, longest common subsequence, and word-based recall. It complements BLEU by focusing on recall rather than precision \cite{9087226}. In our work we calculate ROUGE-1,2, and L values and calculated an average from them. Below is the average value comparison.

\begin{figure}[h]
\centering
\includegraphics[height=4cm, width=0.4\textwidth]{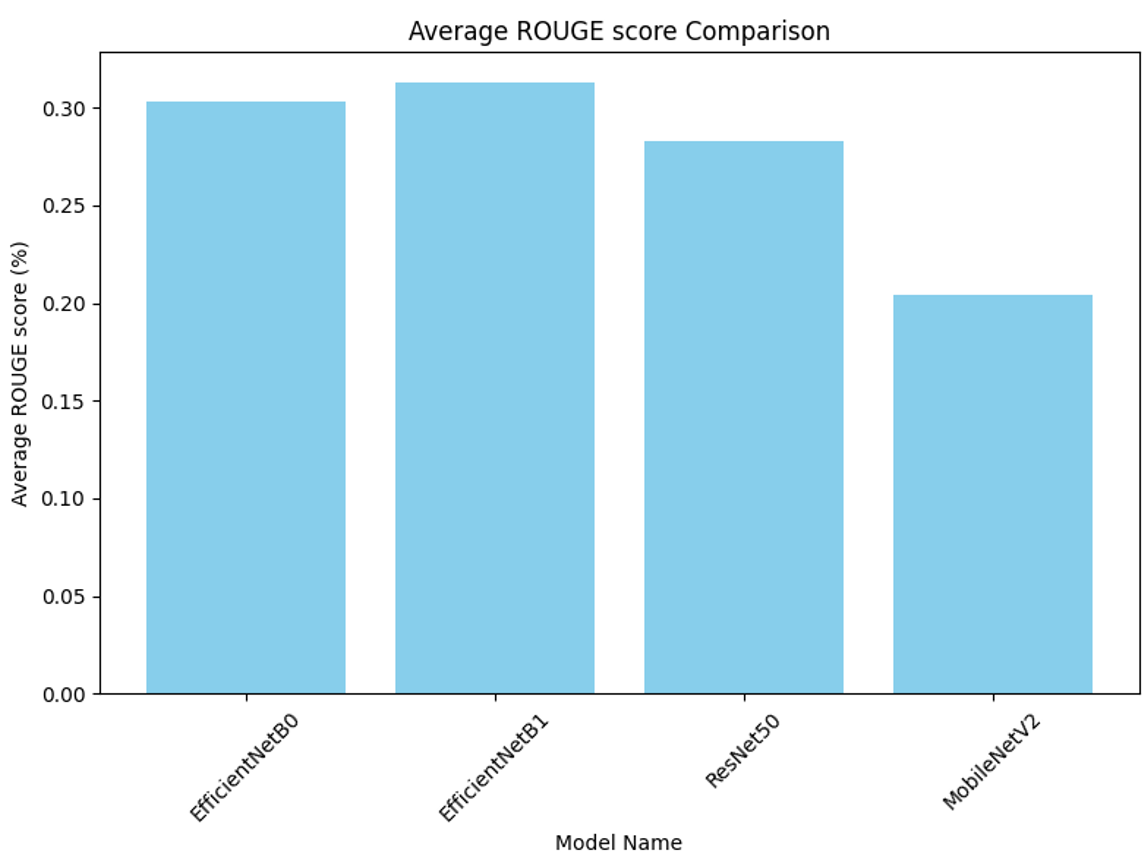}
\caption{Average ROUGE score comparison}
\end{figure}
    
    \item METEOR (Metric for Evaluation of Translation with Explicit Ordering): METEOR considers not only exact word matches but also synonyms, stemmed words, and other linguistic variations through the use of a combination of precision, recall, and alignment-based measures. METEOR scores range from 0 to 1, with higher scores indicating better quality of generated captions. It captures semantic similarity and accounts for different ways of expressing the same idea, making it useful for evaluating the naturalness and fluency of generated captions \cite{9087226}.

\begin{figure}[h]
\centering
\includegraphics[height=4cm, width=0.4\textwidth]{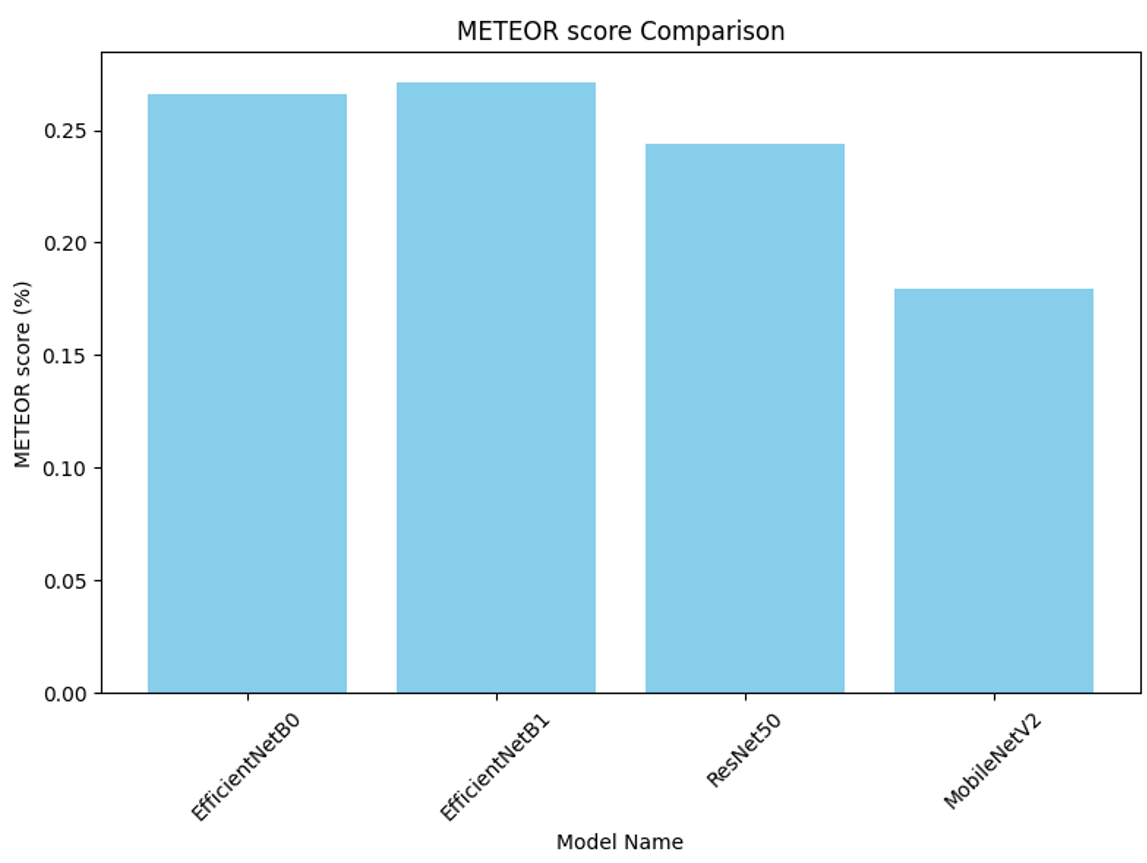}
\caption{METEOR score comparison}
\end{figure}
\end{itemize}

\noindent
Below are some predictions performed on our test data using the best performing model, "EfficientNetB1". The predictions are not exactly accurate, but given the limited hardware resources and less training time, it is acceptable.

\begin{figure}[h]
     \centering
     \begin{subfigure}[t]{0.3\textwidth}
        \centering
        \includegraphics[width=\textwidth, height = 4 cm]{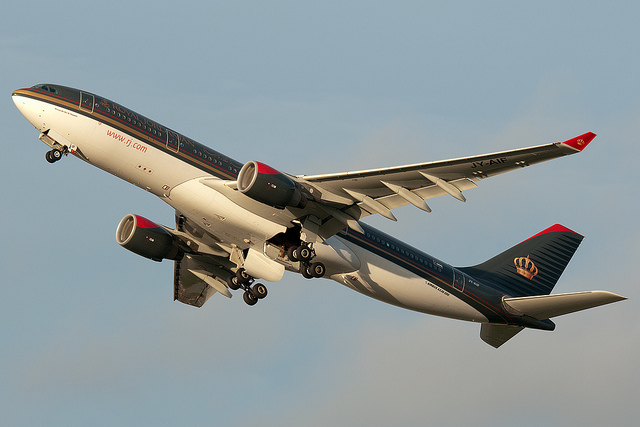}
        \caption*{\small{\textit{"a large white and blue air kite on a blue sky"}}}
    \end{subfigure}
     \hfill
     \begin{subfigure}[t]{0.3\textwidth}
         \centering
         \includegraphics[width=\textwidth, height = 4 cm]{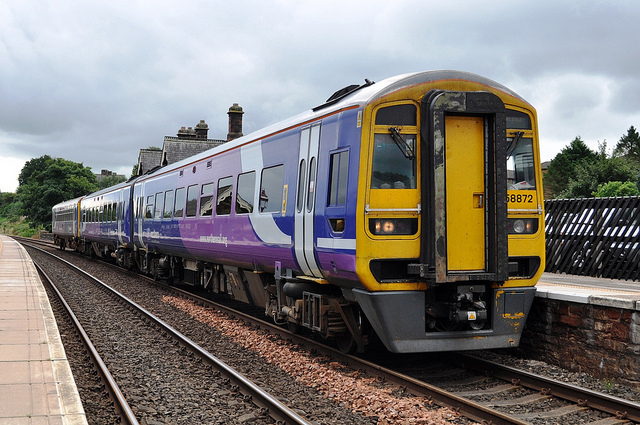}
         \caption*{\small{\textit{"a train is parked on the train tracks"}}}
         \label{}
     \end{subfigure}
     \hfill
     \begin{subfigure}[t]{0.3\textwidth}
         \centering
         \includegraphics[width=\textwidth, height = 4 cm]{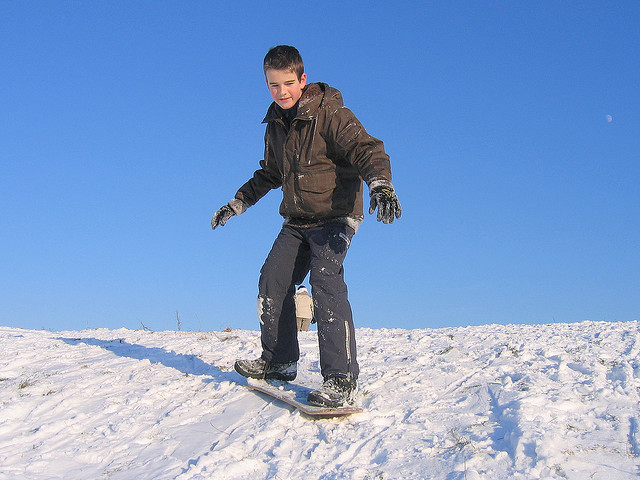}
         \caption*{\small{\textit{"a young boy is on a skateboard in a park"}}}
         \label{}
     \end{subfigure}
        \caption{Prediction on test data}
        \label{fig:three graphs with captions}
\end{figure}


\section{Novelty}
In our proposed methodology, at first we extract the image features by using CNN models before passing it to the encoder model. We have used models like "EfficientNetB0", "EfficientNetB1", "EfficientNetB2", "MobileNetV2", and "ResNet50" etc. But the number of parameters of those models are huge and size of the pretrained weights are also big. Which increases the overall size of our model and training time. In order to use smaller pre-trained CNN model, we have used Frequency Regularization technique proposed by the paper named "Frequency Regularization: Reducing Information Redundancy in Convolutional Neural Networks". We decided to convert the state-of-the-art CNN models to use frequency regularization technique to train on "ImageNet" dataset \cite{5206848}. Then use that pretrained weight to extract features in our architectures. We have used PyTorch library for writing these architectures. Initially, we decided to convert the AlexNet model. Here in this model, we have used the custom Conv2d function from "frereg" package available in the "pip" package manager. This function uses the frequency regularization technique to decrease the parameters. We have also used custom "Linear" function available in the "frereg" package to decrease parameters of the model further.

\begin{algorithm}
\caption{AlexNet Architecture with Frequency Regularization}
\begin{algorithmic}[1]
\STATE \textbf{Define features}:
\STATE $features = $
\STATE \quad $fr.Conv2d(inchannels=3, outchannels=64, kernel\_size=11, stride=4, padding=2)$
\STATE \quad $nn.ReLU(inplace=True)$

\STATE \quad $nn.MaxPool2d(kernel\_size=3, stride=2)$
\STATE \quad $fr.Conv2d(inchannels=64, outchannels=192, kernel\_size=5, padding=2)$
\STATE \quad $nn.ReLU(inplace=True)$
\algstore{myalg1}
\end{algorithmic}
\end{algorithm}

\begin{algorithm}                     
\begin{algorithmic} [1]                   
\algrestore{myalg1}
\STATE \quad $nn.MaxPool2d(kernel\_size=3, stride=2)$
\STATE \quad $fr.Conv2d(in\_channels=192, out\_channels=384, kernel\_size=3, padding=1)$
\STATE \quad $nn.ReLU(inplace=True)$

\STATE \quad $fr.Conv2d(inchannels=384, outchannels=256, kernel\_size=3, padding=1)$
\STATE \quad $nn.ReLU(inplace=True)$
\STATE \quad $fr.Conv2d(inchannels=256, outchannels=256, kernel\_size=3, padding=1)$
\STATE \quad $nn.ReLU(inplace=True)$
\STATE \quad $nn.MaxPool2d(kernel\_size=3, stride=2)$
\STATE \textbf{Define avgpool}:
\STATE $avgpool = nn.AdaptiveAvgPool2d((6, 6))$
\STATE \textbf{Define classifier}:
\STATE $classifier = $
\STATE \quad $nn.Dropout(0.5)$
\STATE \quad $fr.Linear(256 * 6 * 6, 4096)$
\STATE \quad $nn.ReLU(inplace=True)$
\STATE \quad $nn.Dropout(0.5)$
\STATE \quad $fr.Linear(4096, 4096)$
\STATE \quad $nn.ReLU(inplace=True)$
\STATE \quad $fr.Linear(4096, num\_classes)$
\end{algorithmic}
\end{algorithm}

\noindent
We have also converted the "EfficientNetB0" algorithm to use the Frequency Regularization. Here in the Convolution-BatchNorm-ReLU block we have used the custom Conv2d function available in the "frereg" package. Moreover, we employed the custom Conv2d in the "Mobile Inverted Residual Block" too. Finally, we integrated the custom Linear function in the classifer for the "EfficientNetB0" further.   

\begin{algorithm}
\caption{EfficientNetB0 Architecture with Frequency Regularization}
\begin{algorithmic}[1]

\STATE \textbf{Define Convolution-BatchNorm-ReLU block (ConvBNReLU)}
\STATE \quad \textbf{Return:} Sequential(
\STATE \quad \quad fr.Conv2d($in\_channels$, $out\_channels$, $kernel\_size$, stride=$stride$, padding=$padding$, bias=False),
\STATE \quad \quad BatchNorm2d($out\_channels$),
\STATE \quad \quad ReLU6(inplace=True)
\STATE \quad )

\STATE

\STATE

\STATE \textbf{Define Mobile Inverted Residual Block (MBConvBlock)}

\STATE \quad \textbf{Return:} 
\STATE \quad \quad Expand Channels = $in\_channels \times expand\_ratio$
\STATE \quad \quad Use Residual Connection = ($stride == 1$ and $in\_channels == out\_channels$)

\STATE \quad \quad Layers = []
\algstore{myalg2}
\end{algorithmic}
\end{algorithm}

\newpage
\begin{algorithm}                     
\begin{algorithmic} [1]                   
\algrestore{myalg2}

\STATE \quad \quad \quad If $expand\_ratio \neq 1$:

\STATE \quad \quad \quad \quad Append ConvBNReLU($in\_channels$, Expand Channels, kernel\_size=1)

\STATE \quad \quad \quad Append ConvBNReLU(Expand Channels, Expand Channels, $kernel\_size$, stride=$stride$, padding=$kernel\_size // 2$)

\STATE \quad \quad \quad Append fr.Conv2d(Expand Channels, $out\_channels$, kernel\_size=1, bias=False)
\STATE \quad \quad \quad Append BatchNorm2d($out\_channels$)
\STATE \quad \quad self.layers = Sequential(*layers)

\STATE \textbf{Define EfficientNetB0 Model}
\STATE Create stem as ConvBNReLU layer with input channels=3, output channels=32, kernel size=3, stride=2, padding=1
\STATE Create blocks as a sequence of MBConvBlocks:
\STATE \quad - MBConvBlock(32, 16, expand ratio=1, kernel size=3, stride=1)
\STATE \quad - MBConvBlock(16, 24, expand ratio=6, kernel size=3, stride=2)
\STATE \quad - MBConvBlock(24, 24, expand ratio=6, kernel size=3, stride=1)
\STATE \quad - MBConvBlock(24, 40, expand ratio=6, kernel size=5, stride=2)
\STATE \quad - MBConvBlock(40, 40, expand ratio=6, kernel size=5, stride=1)

\STATE \quad  ... (continue with other MBConvBlocks)

\STATE \quad $self.head = ConvBNReLU(320, 1280, kernel_size=1)$
\STATE \quad  $self.avg_pool = nn.AdaptiveAvgPool2d((1, 1))$ \STATE \quad  $self.classifier = fr.Linear(1280, num_classes)$

\end{algorithmic}
\end{algorithm}

The authors have used a technique where they first dropped the parameters with very small value before calculating the total number of parameters. For example, when the total number of parameters of UNet model is checked it showed 31 Millions. But when the total number of parameters is checked after loading the pre-trained model provided by the authors it shows 759 parameters. Although, we have successfully been able to integrate "Frequency regularization" in the "AlexNet" and "EfficientNetB0" architecture and train the model, we could not achieve the satisfactory performance and we were able to drop only 10K parameters from the original models. The accuracy achieved was only 10\%. As the accuracy of this regularized model was not satisfactory, we did not use this for the feature extraction purpose in our work. In the future, we want to improve the accuracy of this regularized model and drop satisfactory amount of parameters before using it in image captioning architecture.

\section{Conclusion \& Future Work}
In conclusion, our project endeavors to harness the potential of image captioning to address real-world challenges and enhance user experiences across various domains. Through meticulous experimentation and performance comparison with pre-trained CNN models, we have gained insights into the nuances of different architectures, paving the way for optimization strategies. Our exploration into integrating frequency regularization techniques to compress widely used CNN models, such as "AlexNet" and "EfficientNetB0", demonstrates our commitment to resource efficiency without compromising effectiveness in generating image captions. Our work signifies the crucial role of CNN models in feature extraction, coupled with the encoder-decoder framework for caption generation. This synergy enables us to develop a robust and efficient image captioning system capable of producing descriptive captions for diverse visual content. Looking ahead, there are several avenues for future research and improvement. One direction involves exploring advanced techniques in feature extraction and model architectures to further enhance the quality and diversity of generated captions. Additionally, incorporating multimodal approaches that integrate information from other modalities, such as audio or text, could enrich the contextual understanding of images and improve caption generation accuracy.

\clearpage	
\bibliographystyle{plain}
\bibliography{MyBib}
\end{document}